\documentclass[runningheads]{llncs}
\usepackage{amsmath, amssymb} 
\usepackage{multirow}
\usepackage{booktabs}
\usepackage{xcolor}
\usepackage{subcaption}
\usepackage{amsmath}
\usepackage{graphicx}
\usepackage{booktabs}
\usepackage{enumitem}

\makeatletter
\def\thanks#1{\protected@xdef\@thanks{\@thanks
        \protect\footnotetext{#1}}}
\makeatother
\title{EVL-MCoT: Enhanced Vision-Language Multi-CoT for Harmful Meme Detection}
 \titlerunning{EVL-MCoT} 
 \author{
Hao Yang \and
Jin Wang\textsuperscript{\dag}\thanks{\textsuperscript{\dag}Corresponding author} \and
Xuejie Zhang
}

\institute{
School of Information Science and Engineering, Yunnan University, Kunming, China \\
\email{yanghao888@stu.ynu.edu.cn, wangjin@ynu.edu.cn, xjzhang@ynu.edu.cn}
}

\begin{document}
%

%
%

%
%

%

\maketitle              

\begin{abstract}

MEMEs are widely used on the internet and often carry strong elements of sarcasm or irony. Understanding their hidden meanings typically requires a joint interpretation of text and vision. Existing methods focus on the dual-stream vision-language model to extract the visual and text simultaneously, which lacks background information and prior knowledge about the comprehensive explanation of MEME. One feasible option is to adopt chain-of-thought (CoT). However, the simple CoT approach lacks multi-perspective thinking, which may compromise the reliability of the resulting answers. Moreover, it often relies on shallow feature fusion, lacking the fusion of local details and fine-grained visual-prompt text alignment. This limitation prevents a deeper understanding of the intricate connections between the visual and the text. Herein, an enhanced vision-language multi-CoT (EVL-MCoT) approach is proposed to address these limitations. By promoting multi-CoT, EVL-MCoT enhances consistency and reduces bias in the decision-making process. Additionally, we design a prototype-guided and context-guided decoding framework, which incorporates visual prototypes to guide the fusion process and enables the model to align textual and visual information more precisely. We achieve promising results on the HatefulMemes and MultiOff datasets. The source code has been publicly released and is available at \url{https://github.com/BGWH123/EVL-MCoT}.


\end{abstract}

\begin{figure}[t]
    \centering
    \includegraphics[width=1.0\textwidth]{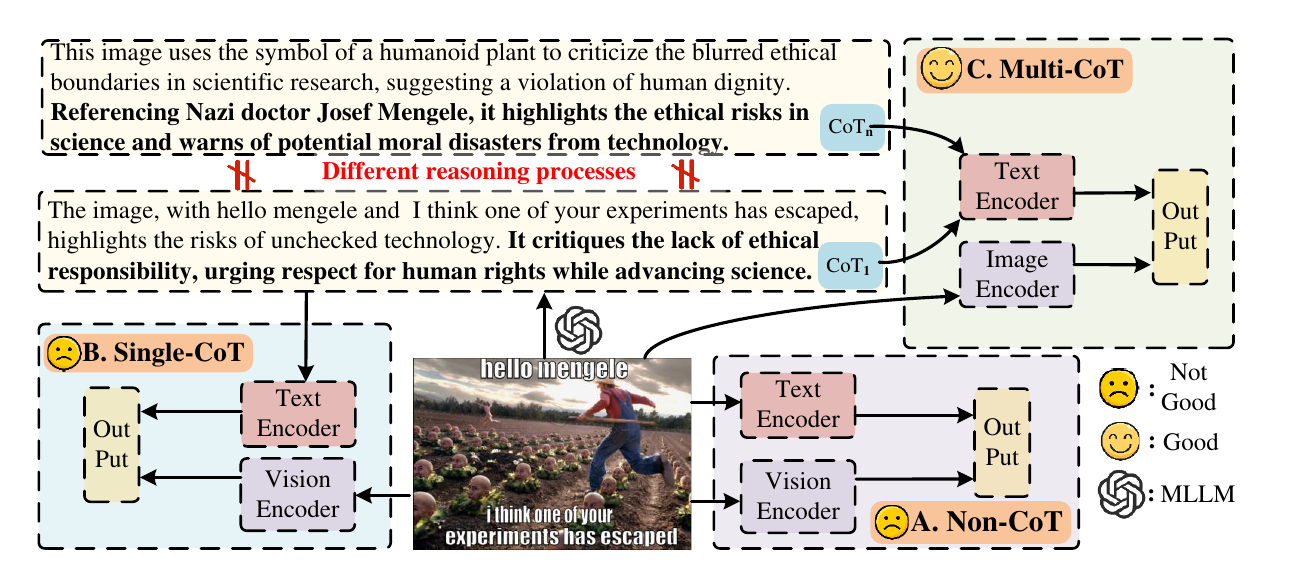}
\caption{\textbf{Non-CoT:} Lacks additional information to support decision-making.
\textbf{Single-CoT:} Includes basic reasoning steps, but follows a single path and therefore lacks alternative perspectives.
\textbf{Multi-CoT:} Provides multiple reasoning chains, which introduces redundancy and helps the model make more robust and accurate decisions.}

    \label{fig1.1}
\end{figure}

\section{Introduction}

Online memes are a unique form of internet culture that combines visual and textual and spread quickly on social media. They shape discourse through humor, irony, and comment but can reinforce stereotypes or promote discrimination, highlighting the need to detect harmful content.

Previous research on harmful meme detection has used primarily two-stream vision language models with task-specific classification layers, which integrate multimodal features learned from text and visual encoders \cite{kiela2020supervisedmultimodalbitransformersclassifying,pramanick-etal-2021-detecting}. However, these models lack the necessary background knowledge and reliable evidence for meme prediction and understanding of deeper semantics and implicit meanings.

Moreover, recent studies have explored using CoT reasoning  \cite{lin2024explainableharmfulmemedetection,kumari2024m3hopcotmisogynousmemeidentification}. Although CoT-based approaches show advantages in enhancing the model's reasoning capabilities, several issues persist. First, their reasoning processes can be unreliable due to their reliance on a single reasoning chain, which makes them susceptible to inconsistencies and errors, as well as being heavily influenced by the prior knowledge from large language models (LLMs)\cite{ma-etal-2023-fedid}. In addition, many methods struggle to capture the fine-grained and detailed elements in visuals\cite{10175600,zheng2024instructiontuningretrievalbasedexamples} that are closely related to the prompt text. This limitation weakens the model's ability to achieve deep alignment and understanding between visual and textual information. The specific issue can be seen in Figure \ref{fig1.1}.

To address these issues, a method called enhanced vision-language multi-CoT (EVL-MCoT) is proposed for harmful meme detection. This approach leverages CoT reasoning to provide the necessary background knowledge and logical foundation for understanding memes. To improve reasoning reliability, multi-CoT are generated, which enhances the consistency of the reasoning process.  Simultaneously, to tackle the challenge of fine-grained fusion, a decoder framework is introduced, consisting of a prototype-guided decoder and a context-guided decoder. The prototype-guided decoder utilizes prototype vectors to progressively steer the integration of visual features. Moreover, the context-guided decoders treat visual prototypes as queries and textual features as key-value pairs, effectively guiding the text encoder to capture visual-contextual information more accurately. Enhanced features derived from hateful and benign texts are fused with optimized visual features. These enriched textual representations are then concatenated and aligned with visual features, enabling the model to better capture and contextually relevant visual elements. Through these mechanisms, EVL-MCoT significantly strengthens the alignment between visual and textual modalities, ultimately enhancing the model’s capability in harmful meme detection.

The main contributions of this study are summarized as follows:
\begin{itemize}[label=\textbullet]
    \item We propose the EVL-MCoT framework, which integrates multi-CoT and advanced multimodal fusion to address unreliable single-chain reasoning and shallow cross-modal alignment in harmful meme detection.
    
    \item EVL-MCoT generates diverse CoT to reduce bias and inconsistencies while enhancing visual-text alignment through prototype decoding and context-guided decoding. 
    
    \item Experiments conducted on these Hatefulmemes \cite{kiela2021hatefulmemeschallengedetecting} and MultiOFF \cite{suryawanshi-etal-2020-multimodal} demonstrate the competitive performance of the proposed method.
\end{itemize}

\section{Related Work}

\noindent\textbf{Hateful Meme Detection.} Object detection (OD)-based models, such as VisualBERT \cite{li2019visualbertsimpleperformantbaseline} and UNITER \cite{henderson2017efficientnaturallanguageresponse}, utilize faster R-CNN detectors \cite{7485869}; however, they suffer from high inference latency. Additionally, contrastive language-image pre-training CLIP-based models, like HateCLIPper \cite{kumar-nandakumar-2022-hate}, offer more efficient end-to-end architectures and have shown improved performance through retrieval-guided contrastive learning. Large LMMs such as Flamingo \cite{alayrac2022flamingovisuallanguagemodel} have demonstrated significant success in meme detection. Flamingo, for instance, achieves superior experimental results, surpassing CLIP-based systems, though it requires expensive fine-tuning. Other models like IDEFICS \cite{laurençon2024mattersbuildingvisionlanguagemodels} and LENS \cite{berrios2023languagemodelsseecomputer} have also shown impressive performance. Research has also explored using LLaVA's zero-shot\cite{luo-etal-2024-zero} prompting capability for hateful meme detection and correction tasks, demonstrating the effectiveness of the pre-trained LLaVA model, albeit with some limitations.

\noindent\textbf{Multi-CoT.}
Recent studies have introduced multi-chain reasoning (MCR) methods to enhance LLMs. Yoran et al. \cite{yoran2024answeringquestionsmetareasoningmultiple,zheng-etal-2024-enhancing} proposed meta-reasoning over multiple chains of thought (MCR), which improves accuracy and interpretability by merging multiple chains through meta-reasoning. Qiu et al. \cite{qiu-etal-2019-dynamically} introduced a dynamically fused graph network (DFGN) to optimize multi-hop reasoning by fusing entity graphs. Nguyen et al. \cite{nguyen2024directevaluationchainofthoughtmultihop} combined CoT reasoning with knowledge graphs to enhance reasoning accuracy. These approaches improve performance in complex tasks.

\noindent\textbf{Visual-Language Models.}
Visual-language models (VLMs)\cite{10736404} like CLIP \cite{radford2021learning} and FLIP \cite{li2023scalinglanguageimagepretrainingmasking} excel in visual recognition tasks. These models employ dual-tower architectures with separate visual and text encoders pre-trained on visual-text datasets. Techniques such as CoOp \cite{Zhou_2022} and CLIP-Adapter \cite{gao2021clipadapterbettervisionlanguagemodels} fine-tune CLIP for new tasks. Recent frameworks like TOP \cite{qu2024riseailanguagepathologists} incorporate linguistic knowledge for better fine-tuning.

\section{Methodology}
\begin{figure}[t]
\centering
\includegraphics[scale=0.41]{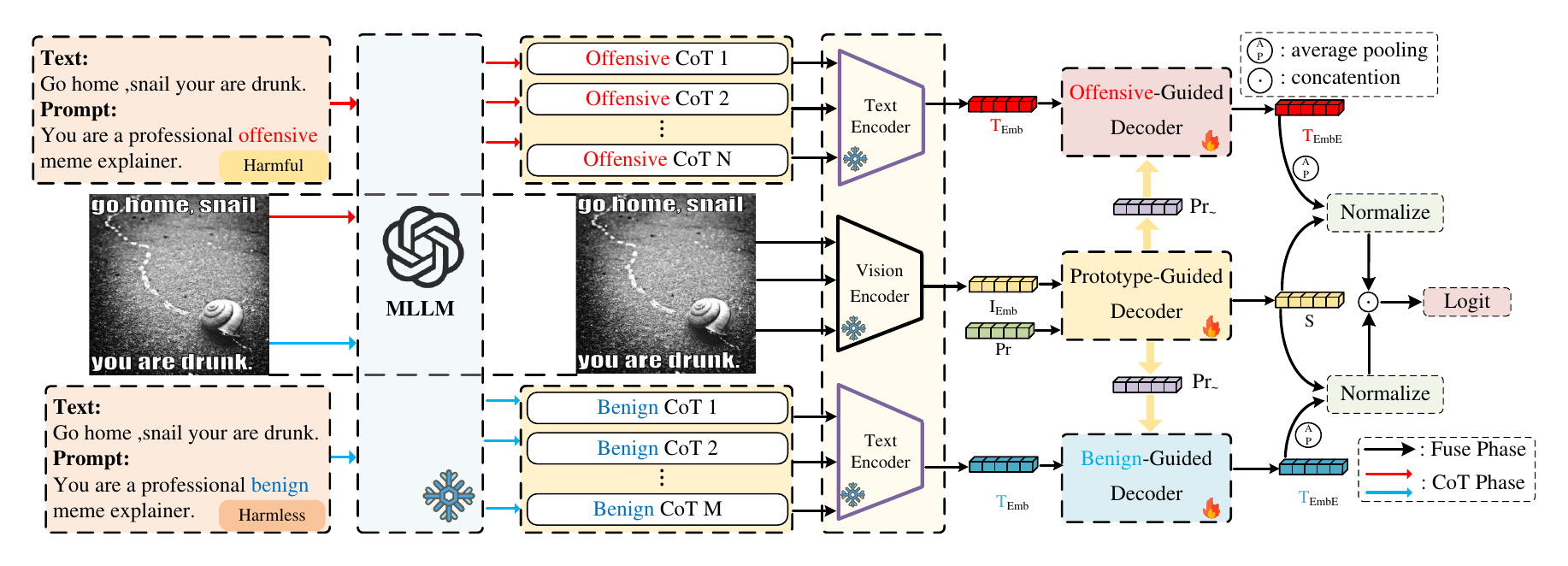}
\caption{
The model architecture comprises a text encoder and a vision encoder, which extract textual and visual features respectively. These features are subsequently fed into multiple decoders, including a prototype-guided decoder, an offensive-guided decoder, and a benign-guided decoder. The model applies normalization and ultimately incorporates average pooling and feature concatenation to compute classification probabilities.} \label{fig3}
\end{figure}

The method framework in Figure \ref{fig3},
 define harmful meme detection as identifying whether a meme contains harmful content. Each meme in the dataset is represented as a tuple \( M = \{ I, T \} \), where \( I \) is the visual and \( T \) is the associated text.

\subsection{Multi-CoT Generation}

In meme prediction with Multimodal Large Language Model (MLLM), the models' extensive knowledge and contextual understanding help identify the necessary background knowledge and reliable evidence to predict memes and understand deeper meanings and implicit implications. However, directly instructing MLLM to perform meme classification tasks may overlook implicit satire, discrimination, or cultural references, leading to misclassification. Moreover, their reasoning processes can be unreliable due to reliance on a single chain of thought and inherent biases from prior knowledge, resulting in inconsistency and errors.

To address this, a multi-CoT is introduced. For each meme example \( M = \{I, T\} \), a prompt \( q^* \) is designed that incorporates a harmfulness indicator \( Z^* \in \{h_0, h_1\} \) where ($h_0$ is benign and $h_1$ is hateful), guiding MLLM (e.g., GPT, LLaVA) to generate reasoning processes \( S^* \) based on the relationship between the image and its corresponding text.

The prompt is formulated as follows:  
\begin{quote}
\emph{You are a professional meme explainer. Please explain why this meme is hateful/benign, using relevant background and commonsense knowledge.}
\end{quote}

To ensure robustness in reasoning, multiple textual interpretations of the meme are considered, covering both harmful and harmless perspectives:  
\emph{\( \text{benign}_t =\{ 1, 2, \dots, m \}\)} and \emph{\( \text{hateful}_t =\{ 1, 2, \dots, n \}\)}  .

By pairing each \(\text{hateful}_t\) with each \(\text{benign}_t\) for the same meme instance, a total of \( n \times m \) combinations are formed, where each pair contains one harmful CoT and one benign CoT. This transforms the problem into a comparative reasoning task, selecting the CoT from each pair that better matches the image as the correct answer.

The model is better equipped to capture nuanced cultural cues, implicit bias, and satirical elements by generating and evaluating these multiple reasoning paths, ultimately improving classification robustness.

\subsection{Prototype-guided Decoder}
We directly extract visual representations using the Long-CLIP\cite{zhang2024longclipunlockinglongtextcapability} model using the frozen visual encoder $E_I(\cdot)$ to process visual inputs effectively. Given an visual $I \in \mathbb{R}^{N \times C \times H \times W}$, where $N$ is the batch size, $C$ represents the number of channels, and $H, W$ denotes the dimensions, the model encodes it into an output tensor:
\begin{equation}
    H = E_I(I) \in \mathbb{R}^{N \times M \times D}
\end{equation}
where $M$ is the number of tokens obtained after encoding, and $D$ is the feature dimension.

A prototype-guided token fusion mechanism is introduced to aggregate these visual tokens into a final representation suitable for similarity computation. Initialize a set of learnable prototype features $Pr \in \mathbb{R}^{P \times D}$, where D is the feature dimension, $P$ is the number of prototype features. Specifically, we apply a cross-attention layer where the prototypes $Pr$ serve as queries $Q$, and the extracted visual tokens $H$ serve as keys $K$ and values $V$:
\setlength{\abovedisplayskip}{4pt}  
\setlength{\belowdisplayskip}{4pt}  
\begin{equation}
    Pr_{{\sim}} = \text{Norm}(\text{Softmax}(\frac{Pr\ H^\top}{\sqrt{D}}) H) + Pr
\end{equation}
\noindent where $\text{Norm}(\cdot)$ denotes layer normalization, and $\text{Softmax}(\cdot)$ ensures an adaptive weighting mechanism for feature aggregation. This process enables semantically similar tokens to be grouped under the same prototype, allowing each prototype to capture more global contextual information.

Finally, to derive the  visual features $S$, we employ an attention-based feature fusion approach:

\setlength{\abovedisplayskip}{4pt}  
\setlength{\belowdisplayskip}{4pt}  
\begin{equation}
Pr_{l,i}^t = W_a Pr_{\sim l,i}.
\end{equation}
\begin{equation}
A_{l,i} = \frac{\exp(W_b^\top \tanh(W_v Pr_{l,i}^t))}{\sum_{j=1}^{N_l} \exp(W_b^\top \tanh(W_v Pr_{l,j}^t))}.
\end{equation}
\begin{equation}
S = W_c \sum_{i=1}^{N} A_{l,i} Pr_{l,i}.
\end{equation}

\noindent where $W_a, W_c, W_v \in \mathbb{R}^{D \times D}$ and $W_b \in \mathbb{R}^{D \times 1}$ are trainable weight matrices. The function $\tanh(\cdot)$ provides non-linearity, while $A$ represents the learned attention weights for each prototype in the final visual features $S$.

\begin{figure}[t]
    \centering
    \includegraphics[width=0.95\textwidth]{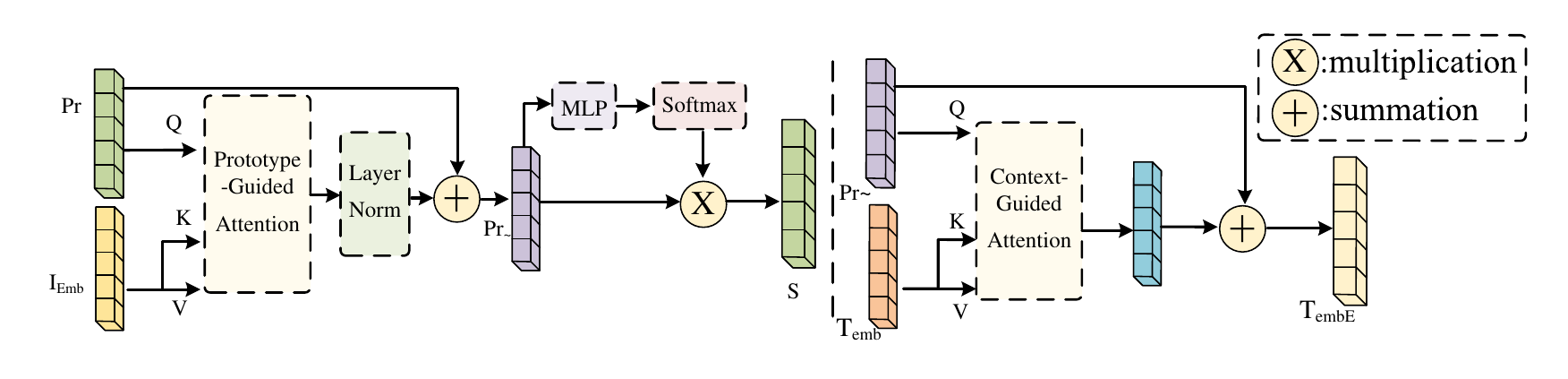}
    \caption{\textbf{Left:} prototype-guided decoder. \textbf{Right:} context-guided decoder.}
    \label{fig3.2}
\end{figure}

Framework in Figure \ref{fig3.2}. This fusion method enhances the ability of the model to capture long-range dependencies and contextual relationships within visual, ultimately improving its capability in multimodal understanding.

\subsection{Context-guided Decoder}
For the text part, we first encode both the \textit{hateful text} and \textit{benign text} using the frozen Long-CLIP \cite{zhang2024longclipunlockinglongtextcapability} text encoder, obtaining the corresponding text feature representations \( T_{emb} \). To further enhance these representations, we integrate visual context information, including local patch representations and global prototype representations. By incorporating visual priors into the textual encoding, the model effectively bridges the gap between vision and text, thereby improving cross-modal alignment.

To achieve this, we employ a context-guided attention layer, where the visual prototype features \( Pr_{\sim} \) serve as the queries \( Q \), while the text features \( T_{emb} \) serve as both the keys \ K \ and values \( V \). The attention mechanism is defined as follows:
\setlength{\abovedisplayskip}{4pt}  
\setlength{\belowdisplayskip}{4pt}  
\begin{equation}
    T_{\text{embE}} = \text{Softmax} \left( \frac{Pr_\sim \ T_{\text{emb}}^\top}{\sqrt{D}} \right) T_{\text{emb}} + Pr_\sim
\end{equation}

Framework in Figure \ref{fig3.2}. By leveraging learnable visual prototypes as attention queries, our model enables fine-grained semantic refinement of textual representations, enriching text embeddings with visual context.

\begin{table}[t]
\caption{Comparison of results (\%) on the HatefulMeme dataset. Bold indicates the best results. The proposed EVL-MCoT is evaluated with the setting n=3, m=3 using LLaVA as the MLLM. \textbf{Note:} \emph{OCR:} Optical Character Recognition; \emph{LLaVA:} Language-Visual-Audio; \emph{MOMENTA:} is a multimodal framework for detecting harmful memes and their targets.}
\centering
\small
\setlength{\tabcolsep}{3pt}
\renewcommand{\arraystretch}{0.7} 
\begin{tabular}{llcccc}
\toprule
Type & Model & \multicolumn{2}{c}{TestSeen} & \multicolumn{2}{c}{TestUnseen} \\
     &       & Acc. & AUROC & Acc. & AUROC \\
\midrule
Unimodal & Image-Grid & 52.00 & 52.63 & - & - \\
         & Image-Region & 52.13 & 55.92 & 60.28 & 54.64 \\
         & TextBERT & 59.20 & 65.08 & 63.60 & 62.65 \\
\midrule
Multimodal & LateFusion & 59.66 & 64.75 & 64.06 & 64.44 \\
           & ConcatBERT & 59.13 & 65.79 & 65.90 & 66.28 \\
           & MMBT-Grid & 60.06 & 67.92 & 66.85 & 67.24 \\
           & MMBT-Region & 60.23 & 70.73 & 70.10 & 72.21 \\
           & ViLBERT & 62.30 & 70.45 & 70.86 & \underline{73.39} \\
           & VisualBERT & 63.20 & 71.33 & \underline{71.30} & 73.23 \\
           & ViLBERTCC & 61.10 & 70.03 & 70.03 & 72.78 \\
           & VisualBERTCOCO & \underline{64.73} & \underline{71.41} & 69.95 & 74.95 \\
\midrule
MultimodalLLM & Flamingo-80B & - & - & - & 70.00 \\
              & IDEFICS-80B & - & - & - & 60.60 \\
              & LLaVA-Llama-2-13B & 63.00 & 65.77 & 62.15 & 63.92 \\
              & GPT-4o & - & - & 65.00 & - \\
              & Mod-Hate & - & - & 58.00 & 64.50 \\
              & MQwen2VL-2B & - & - & 59.70 & 64.10 \\
              & MQwen2VL-7B & - & - & 64.10 & 71.10 \\
              & LLaVA-1.5-7B & - & - & 66.90 & 69.90 \\
              & Flamingo-9B+OCR & - & 57.30 & - & - \\
              & MOMENTA & - & - & 69.20 & 61.30 \\
\midrule
Other & EVL-MCoT & \textbf{75.88} & \textbf{79.25} & \textbf{75.57} & \textbf{79.40} \\
\bottomrule
\end{tabular}

\label{tab4}
\end{table}

\subsection{Training Strategy}

After obtaining the enhanced hateful and benign text features $T^h_{embE}$ and $T^b_{embE}$, along with the refined visual features $S$, we compute their alignment score as follows:
\setlength{\abovedisplayskip}{4pt}  
\setlength{\belowdisplayskip}{6pt}  
\begin{equation}
L = \text{CE} \left( \tau \cdot \frac{\text{Concat}(\text{AvgPool}(T^h_{embE}), \text{AvgPool}(T^b_{embE})) \cdot S^\top}{|\text{Concat}(\text{AvgPool}(T^h_{embE}), \text{AvgPool}(T^b_{embE}))|_2 |S|_2}, \text{GT} \right)
\end{equation}

\noindent where $\text{AvgPool}(\cdot)$ extracts global contextual information, $\text{Concat}(\cdot)$ concatenates hateful and benign text features, $\tau$ is a learnable temperature parameter, and $\text{GT}$ represents the ground truth labels.

\begin{table}[t]
\caption{Comparison of results (\%) on the MultiOFF dataset. Bold indicates the best results. The proposed EVL-MCoT is evaluated with the setting n=3, m=3 using GPT-4 as the MLLM. 
\textbf{Note:} \emph{FT:} Fine-Tuning; \emph{RT:} Random Training; \emph{RGCL:} Relational Graph Contrastive Learning; \emph{SFT:} Supervised Fine-Tuning.}
\centering
\small
\setlength{\tabcolsep}{3pt}
\renewcommand{\arraystretch}{0.7} 

\begin{tabular}{llcc}
\toprule
Type & Model & Acc. & \textbf{$F_1$} \\
\midrule
\multirow{8}{*}{Unimodal} 
         & ResNet50 (FT) & 63.7 & \underline{62.3} \\
         & ResNet50 (RT) & 55.7 & 56.2 \\
         & ViT & 62.4 & 55.9 \\
         & BiLSTM & 60.4 & 60.6 \\
         & BiLSTM + Attention & 59.7 & 57.8 \\
         & BERT & 61.7 & 61.0 \\
         & m-BERT & 61.1 & 57.4 \\
         & XLM-R & 63.0 & 58.0 \\
\midrule
\multirow{5}{*}{Multimodal} 
         & Late Fusion & \underline{65.7} & 56.8 \\
         & Attentive Fusion & 62.4 & 53.8 \\
         & VisualBERT COCO & 68.9 & 50.3 \\
         & CLIP & 65.1 & 60.1 \\
         & ALBEF & 61.7 & 61.3 \\
\midrule
\multirow{6}{*}{SFT} 
         & LLaVA-1.5-7B (SFT + zero-shot) & 62.8 & 32.5 \\
         & LLaVA-1.5-7B (SFT + few-shot) & 56.0 & 38.9 \\
         & Qwen2VL-2B (SFT + zero-shot) & 62.4 & 51.7 \\
         & Qwen2VL-2B (SFT + few-shot) & 62.3 & 29.3 \\
         & Qwen2VL-7B (SFT + zero-shot) & 63.1 & 29.7 \\
         & Qwen2VL-7B (SFT + few-shot) & 62.0 & 53.7 \\
\midrule
\multirow{3}{*}{Other} 
         & GPT-4o & 63.1 & 15.4 \\
         & RGCL & 53.7 & 45.1 \\
         & EVL-MCoT & \textbf{70.0} & \textbf{63.8} \\
\bottomrule
\end{tabular}

\label{tab:multioff_results}
\end{table}

\section{Experiments}
\subsection{Experimental Setup}
\textbf{Datasets.} Two publicly available meme datasets are used for evaluation: Hateful Memes \cite{kiela2021hatefulmemeschallengedetecting} and MultiOFF \cite{suryawanshi-etal-2020-multimodal}. The Hateful Memes dataset contains over 10,000 multimodal examples for detecting hate speech in memes, labeled as \emph{hateful} or \emph{non-hateful}. Its test set is divided into \emph{test\_seen} and \emph{test\_unseen}, where \emph{test\_seen} includes samples that share similar distributions with the training data, and \emph{test\_unseen} consists of samples with novel image-text combinations. The MultiOFF dataset consists of 743 memes, annotated as \emph{offensive} or \emph{non-offensive}.

\noindent\textbf{Baseline.}
In this study, we compare our model with several harmful meme detection baselines, including 1) \textbf{Unimodal}, 2) \textbf{Multimodal}, 3) \textbf{MultimodalLLM}, and 4) \textbf{Other}. For evaluation, the Hateful Memes dataset is assessed using area under the curve \textbf{(AUC)} and accuracy \textbf{(ACC)}, while the MultiOFF dataset employs \textbf{$F_1$}-score and \textbf{ACC} as its evaluation metrics.
\begin{figure}[htbp]
    \centering
    \begin{subfigure}[b]{0.3\textwidth}
        \centering
        \includegraphics[width=\textwidth]{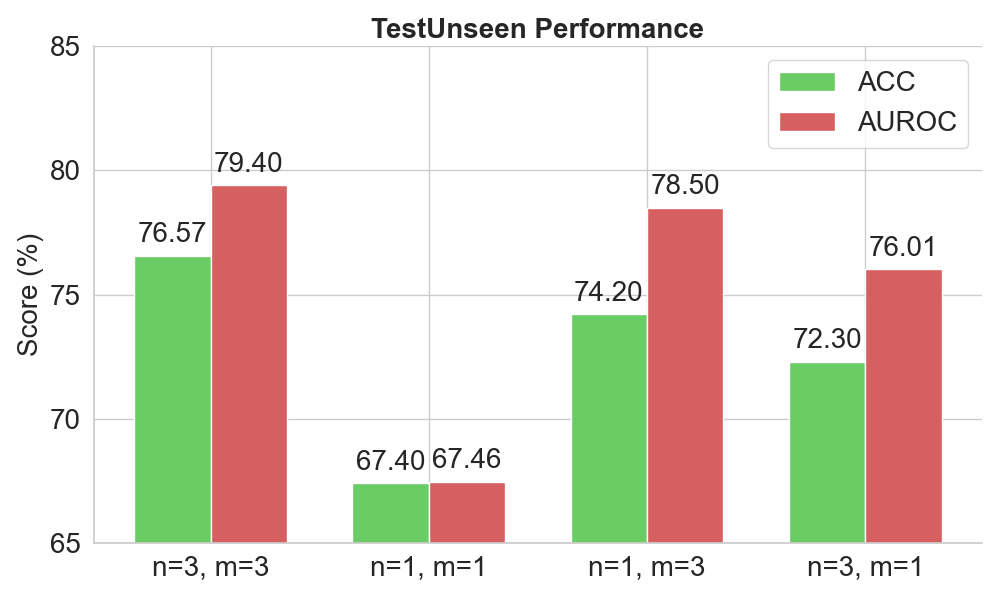}
        \caption{Effect of the Number of CoTs on Test Seen Performance}
        \label{fig:seen}
    \end{subfigure}
    \hfill
    \begin{subfigure}[b]{0.3\textwidth}
        \centering
        \includegraphics[width=\textwidth]{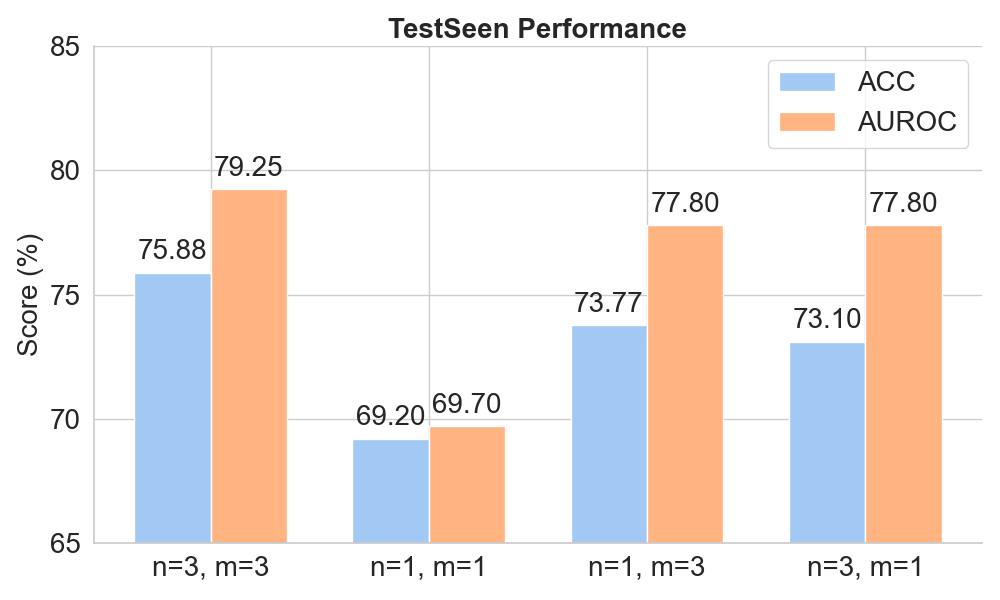}
        \caption{Effect of the Number of CoTs on Test UnSeen Performance}
        \label{fig:unseen}
    \end{subfigure}
    \hfill
    \begin{subfigure}[b]{0.3\textwidth}
        \centering
        \includegraphics[width=\textwidth]{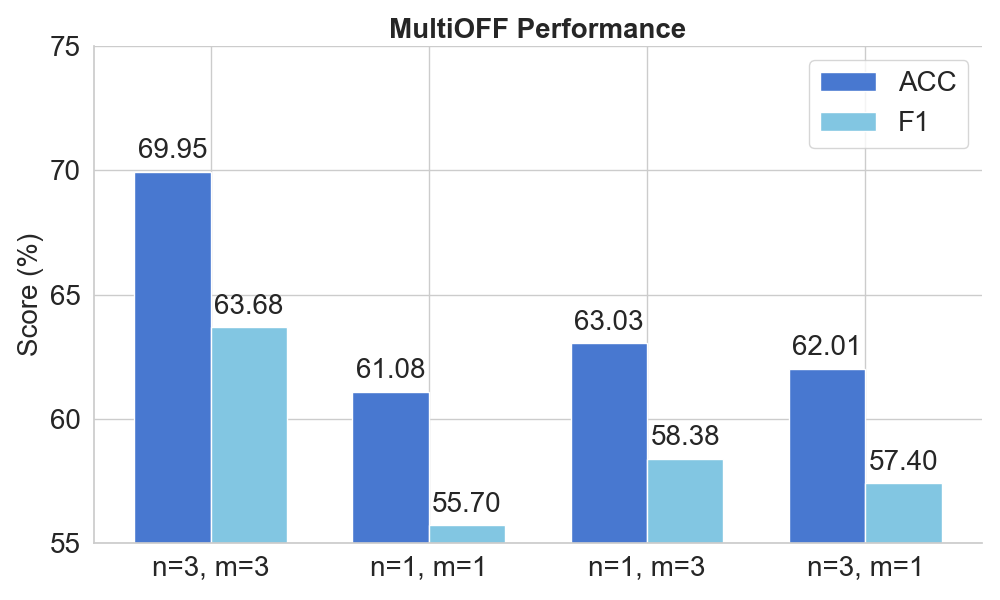}
        \caption{Effect of the Number of CoTs on MultiOFF Performance}
        \label{fig:multioff}
    \end{subfigure}

    \vskip\baselineskip

    \begin{subfigure}[b]{0.45\textwidth}
        \centering
        \includegraphics[width=0.8\textwidth]{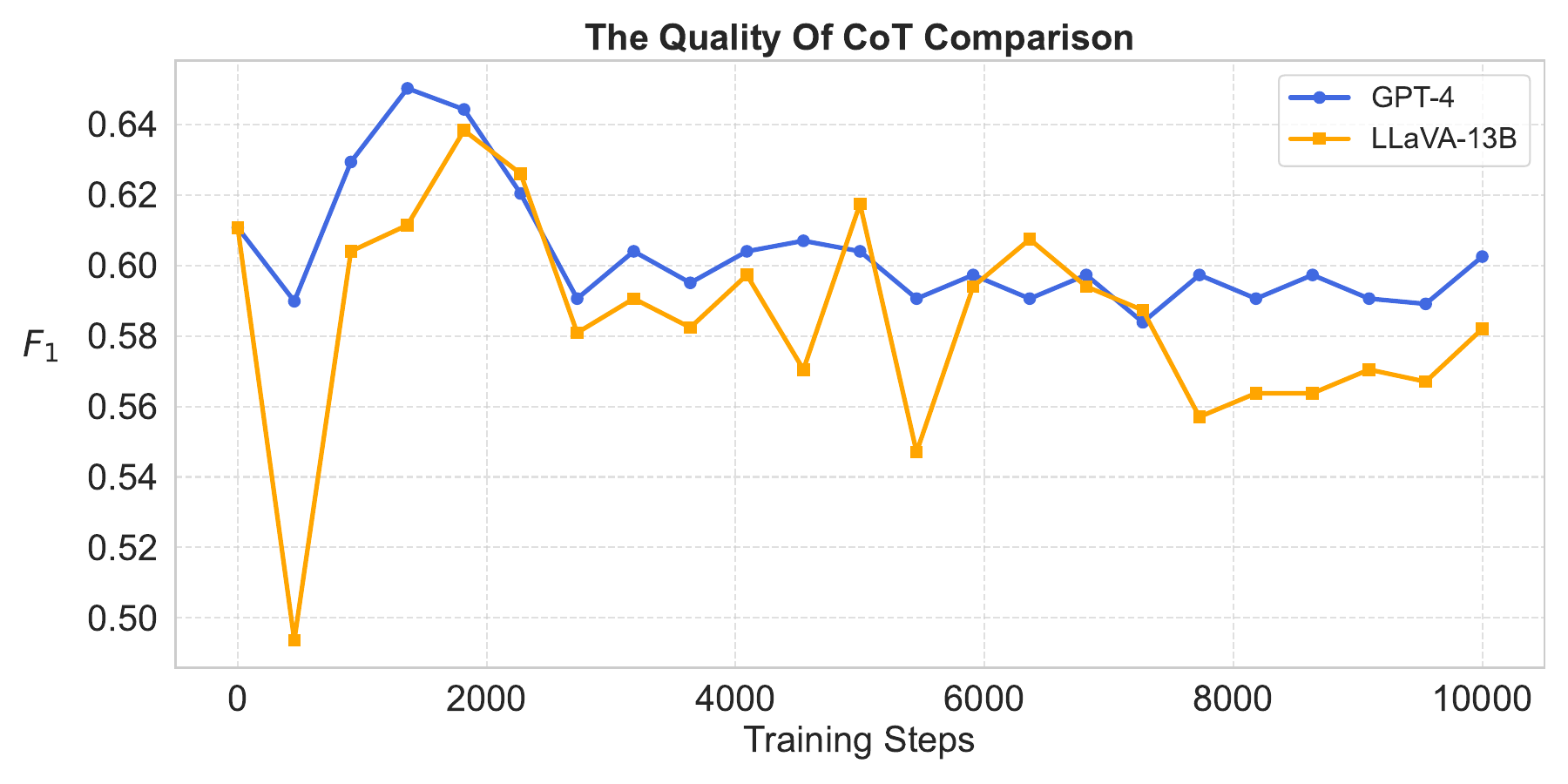}
        \caption{Different MLLM-Generated CoT in MultiOFF Performance}
        \label{fig:cot}
    \end{subfigure}
    \hfill
    \begin{subfigure}[b]{0.45\textwidth}
        \centering
        \includegraphics[width=0.8\textwidth]{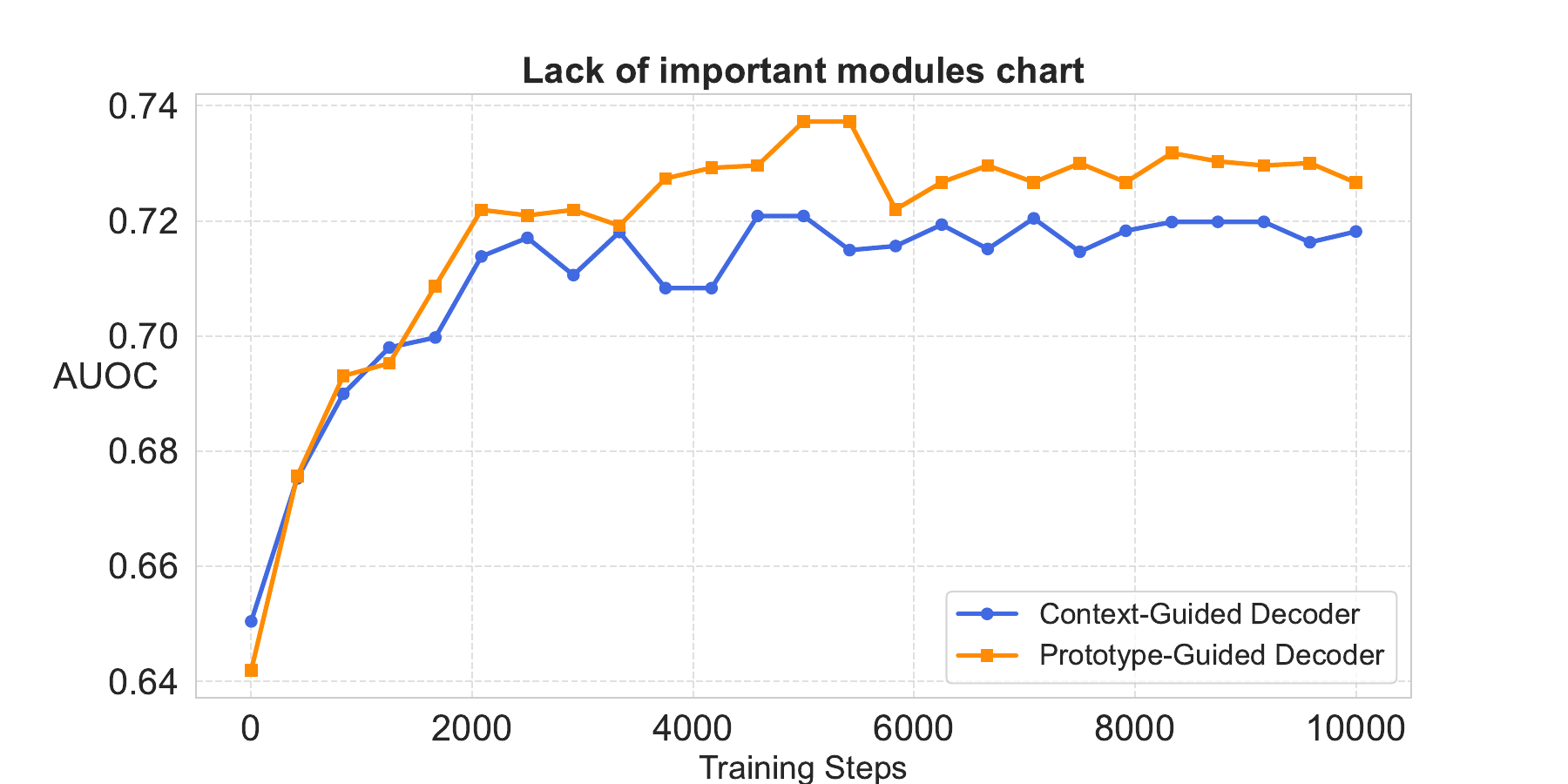}
        \caption{Lack of Importance in MultiOFF Performance}
        \label{fig:final}
    \end{subfigure}

    \caption{Multi-CoT and Module Analysis.}

    \label{fig:cotfig}  
\end{figure}

\begin{table}[t]
\small
\setlength{\tabcolsep}{3pt}
\renewcommand{\arraystretch}{0.7} 
\centering
\caption{Model Performance Comparison on Multimodal Datasets}
\label{tab:performance}
\begin{tabular}{lcccccc}
\toprule

\textbf{Dataset} & \multicolumn{4}{c}{\textbf{HatefulMeme}} & \multicolumn{2}{c}{\textbf{MultiOFF}} \\
\cmidrule(lr){2-5} \cmidrule(lr){6-7}
\ & \multicolumn{2}{c}{\textbf{TestSeen}} & \multicolumn{2}{c}{\textbf{TestUnseen}} & \multicolumn{2}{c}{\textbf{Test}} \\
 \textbf{Model}& \textbf{ACC} & \textbf{AUROC} & \textbf{ACC} & \textbf{AUROC} & \textbf{ACC} & \textbf{$F_1$} \\
\midrule
EVL-MCoT & 75.88 & 79.25 & 76.57 & 79.40 & 69.95 & 63.68 \\
w/o Long-CLIP & 68.68 & 71.77 & 68.23 & 72.63 & 62.60 & 59.19 \\
w/o Prototype-Guided Decoder & 67.85 & 75.61 & 68.23 & 75.60 & 62.60 & 59.19 \\
w/o Context-Guided Decoder & 63.82 & 64.22 & 63.90 & 59.06& 64.83 & 58.39 \\
w/o CoT ($n{=}1,m{=}1$) & 69.20 & 69.70 & 67.40 & 67.46 & 61.08 & 55.70 \\
w/o CoT ($n{=}1,m{=}3$) & 73.77 & 77.80 & 74.20 & 78.50 & 63.08 & 58.38 \\
w/o CoT ($n{=}3,m{=}1$) & 73.10 & 77.80 & 72.30 & 76.01 & 62.01 &  57.40\\
\bottomrule
\label{tab:Ablative Studies}
\end{tabular}

\smallskip
\footnotesize 
\end{table}

\noindent\textbf{Model Selection.}
The standard CLIP model struggles with inputs exceeding 20 tokens, while our CoT approach requires processing 150-word sequences. To address this, we use Long-CLIP \cite{zhang2024longclipunlockinglongtextcapability}, which extends the context length to 248 tokens, enabling more effective reasoning over longer texts.

\noindent\textbf{Implementation Details.}
For details, the batch size is set to 8 with gradient accumulation every 2 steps. The optimizer is set to AdamW \cite{loshchilov2019decoupledweightdecayregularization}, with a learning rate of 5e-5, a weight decay of 0.01, and a cosine learning rate scheduler. All experiments were conducted on NVIDIA GeForce RTX 3090 and 4090 GPUs.

\subsection{Results}
\textbf{Hatefulmeme.} The experiment results in Table \ref{tab4} show that EVL-MCoT achieves the best performance on both TestSeen and TestUnseen datasets. Specifically, it reaches an accuracy of 75.88 and AUROC of 79.25 on TestSeen and 75.57 accuracy with an AUROC of 79.40 on TestUnseen. This significant performance improvement demonstrates the superior capability of EVL-MCoT in cross-modal understanding and reasoning, particularly in generalization to unseen data.

\noindent\textbf{MultiOFF.} The experimental results in Table \ref{tab:multioff_results} show that EVL-MCoT achieves the best performance on the MultiOFF dataset, with an accuracy of 70.0 and an \textbf{$F_1$} score of 63.8. EVL-MCoT outperforms both unimodal and multimodal baselines.

\subsection{Analysis}
\noindent\textbf{Result Analysis.} The multi-CoT approach enhances model robustness by generating multiple reasoning chains, with an MLLM simulating both supportive and opposing arguments. This decomposition of meme semantics improves contextual understanding and reduces randomness, leading to more accurate and consistent interpretations. From a multimodal perspective, the framework integrates prototype-guided decoding and context-aware attention for better visual-text alignment. The prototype-guided decoder provides structured references for multimodal content, while context-aware attention highlights key semantic components.

\noindent\textbf{CoT Number and Quality Analysis.} As shown in Figure~\ref{fig:cotfig}(\subref{fig:seen}), Figure~\ref{fig:cotfig}(\subref{fig:unseen}), and Figure~\ref{fig:cotfig}(\subref{fig:multioff}), the number and quality of CoTs significantly impact task performance. A small number of CoTs limits the model's reasoning ability and accuracy, leading to unstable performance. However, more CoTs expose the model to diverse reasoning paths, enhancing reasoning and generalization. High-quality CoT improve performance by providing more meaningful reasoning paths, as seen in Figure~\ref{fig:cotfig}(\subref{fig:cot}).

\noindent\textbf{Prototype and Context-Guided Decoder Analysis.} Figure~\ref{fig:cotfig}(\subref{fig:final}) shows that the context-guided decoder is critical in helping the model understand complex semantics and integrate multimodal information. While prototype-guided decoder offers useful structural priors, its absence of contextual information limits the model's adaptability.

\begin{figure}[t]
\centering

\includegraphics[scale=0.5]{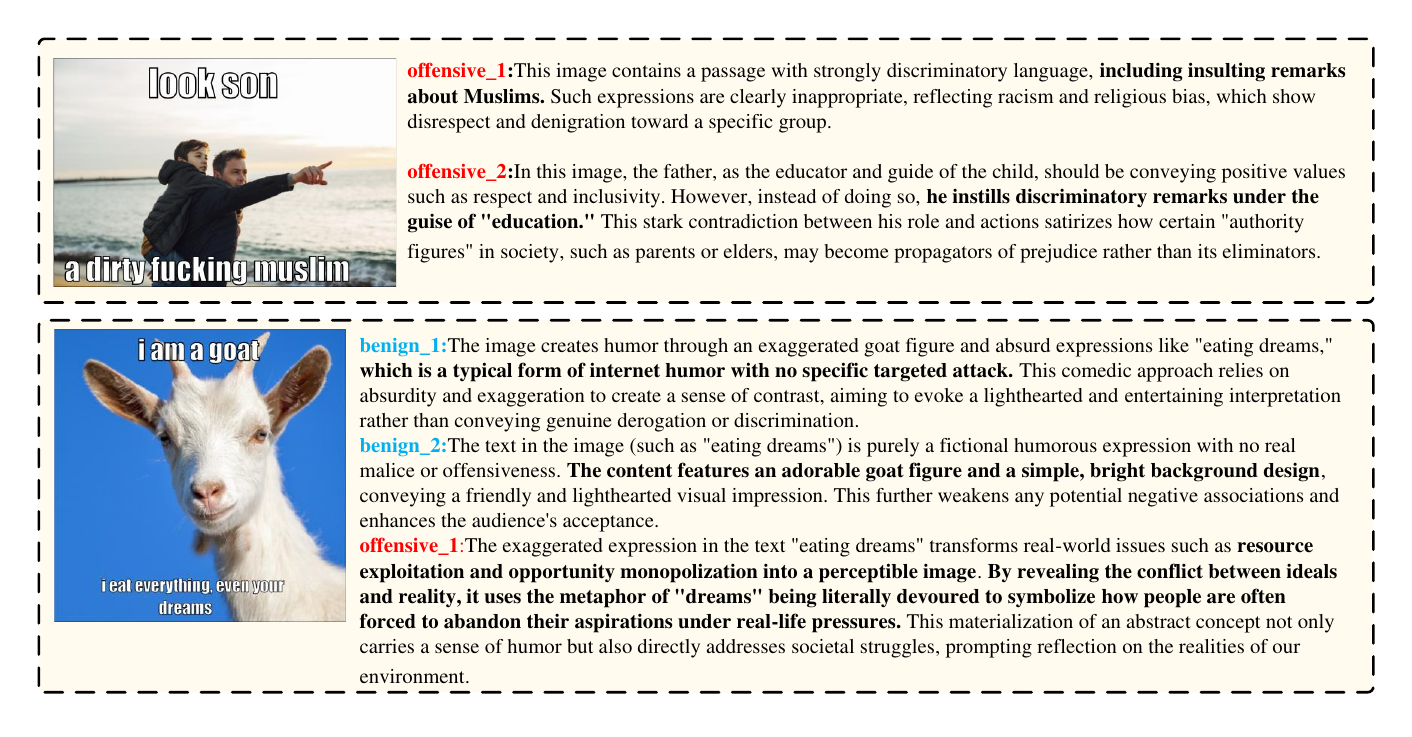}
\caption{Two examples for generating GPT's Multi-CoT. 
} \label{fig4}
\end{figure}
\subsection{Ablation Studies}
The ablation study results in Table~\ref{tab:Ablative Studies} demonstrate the critical role of each component in EVL-MCoT for multimodal hate speech detection. Experiments on HatefulMemes and MultiOFF show that removing Long-CLIP significantly reduces ACC and AUROC, underscoring the importance of long-context multimodal representations. Similarly, removing the prototype-guided decoder leads to comparable performance degradation, highlighting its role in structured feature learning. Excluding the context-guided decoder also lowers performance, confirming its effectiveness in enhancing contextual alignment.

Removing CoT reasoning entirely (\(n{=}1, m{=}1\)) causes a substantial drop in accuracy, validating the impact of structured reasoning. Reducing either reasoning paths or steps (\(n{=}1, m{=}3\) or \(n{=}3, m{=}1\)) leads to smaller declines, indicating that both multiple reasoning paths and multi-step reasoning are necessary for optimal performance. In summary, Long-CLIP and prototype-guided decoder are key to effective multimodal feature extraction, while context-guided decoder enhances alignment, and CoT reasoning boosts overall reasoning ability and robustness.

\subsection{CoT-Explainability}

The CoT explanations generated by MLLM demonstrate structured reasoning but often emphasize different aspects of the same text, leading to diverse interpretations. The variability in MLLM-generated CoT reasoning paths when processing hateful and benign content is illustrated in Figure \ref{fig4}.

MLLM analysis often generates different CoTs depending on which aspect image, text, or implied meaning is emphasized. Even when the final judgment remains the same, the reasoning paths can vary significantly. This diversity highlights the interpretability and context sensitivity of large language models. While such flexibility enables understanding, it also poses challenges for consistency in automated moderation. Embracing multi-perspective reasoning may therefore enhance the robustness and fairness of moderation systems.

\section{Conclusion}
The EVL-MCoT framework proposed in this study addresses the limitations of existing methods through multi-chain reasoning and a cross-modal enhanced decoder, significantly improving harmful meme detection. Experimental results show that EVL-MCoT performs excellently on the HatefulMemes and MultiOFF datasets, and ablation studies confirm the necessity of each component. The generated CoT explanations enhance the model's interpretability. Future work aims to explore strategies for shortening the reasoning chains in EVL-MCoT while maintaining performance, and to extend its application to other multimodal tasks.

\section*{Acknowledgements}
This work was supported by the National Natural Science Foundation of
China (NSFC) under Grant Nos. 61966038 and 62266051, and the Postgraduate
Research and Innovation Foundation of Yunnan University, China under
Grant No. KC-24248816.

\bibliographystyle{splncs04}
\bibliography{refer}
\end{document}